\documentclass[runningheads]{llncs}

\usepackage{eccv}

\usepackage{eccvabbrv}

\usepackage{graphicx}
\usepackage{booktabs}

\usepackage{multirow}

\usepackage[accsupp]{axessibility}  

\usepackage{hyperref}

\hypersetup{
    colorlinks=true,
    linkcolor=blue,
    urlcolor=blue,
    citecolor=blue
}

\usepackage{orcidlink}

\begin{document}

\title{Wan-Dancer: A Hierarchical Framework for Minute-scale Coherent Music-to-Dance Generation} 

\titlerunning{Wan-Dancer}


\author{
  Mingyang Huang,
  Peng Zhang,
  Li Hu,
  Guangyuan Wang,
  Ruoshi Zhang,
  Yi Lu,
  Gang Cheng,
  Bang Zhang
}

\authorrunning{Mingyang, H., Peng, Z., Li, H., Guangyuan, W., Bang, Z.}


\institute{Tongyi Lab, Alibaba Group\\
\email{\{hongcan.hmy, futian.zp, hooks.hl, yixuan.wgy, ruoshi.zrs, ly449033, chenggang.cg, zhangbang.zb\}@alibaba-inc.com}}

\maketitle

{
    \begin{center}
        \textbf{\href{https://humanaigc.github.io/wan-dancer-project/}{Project}} | 
        \href{https://github.com/Wan-Video/Wan-Dancer}{GitHub} |
        \href{https://modelscope.ai/studios/Wan-AI/Wan-Dancer}{MS Space} |
        \href{https://www.modelscope.cn/models/Wan-AI/Wan-Dancer-14B}{MS Model} |
        \href{https://huggingface.co/Wan-AI/Wan-Dancer-14B}{HF} |
        \href{https://arxiv.org/abs/2607.09581}{Paper}
    \end{center}
}

\begin{abstract}
  Generating long-duration, high-definition, and rhythmically synchronized dance videos directly from music remains a significant challenge, primarily due to the temporal constraints of current diffusion models, which typically fail beyond 20 seconds. Existing approaches, whether they rely on intermediate 3D skeletons or on end-to-end video synthesis, suffer from temporal drift, identity inconsistency, and repetitive motion patterns when extended to longer horizons. To address these limitations, we propose a novel hierarchical framework for minute-scale coherent music-to-dance generation. Our method decouples the process into global keyframe planning and local temporal refinement, leveraging full-track musical context to ensure long-range coherence. Key innovations include dynamic frame rate adaptation via time-mapped RoPE embeddings for precise alignment, an optical-flow-based loss function to enhance motion continuity, and motion-speed control to preserve high-fidelity details during rapid movements. Extensive experiments demonstrate that our framework surpasses the conventional duration barrier, generating stable, 720p/30fps videos exceeding one minute with superior temporal stability. Furthermore, the model exhibits robust versatility across five distinct dance genres, conditioned on both audio and textual prompts, establishing a new state-of-the-art in coherent, long-form dance video synthesis.
\end{abstract}

\section{Introduction}
\label{sec:introduction}

Recent advances in diffusion-based video generation have significantly elevated the visual fidelity and motion realism of short-form content. However, current state-of-the-art models, such as Wan~\cite{wan2025wan}, HunyuanVideo~\cite{wu2025hunyuanvideo}, and Seedance~\cite{seedance2025seedance}, remain fundamentally constrained to temporal windows of 5-15 seconds. This limitation stems from the quadratic computational complexity of self-attention mechanisms over long sequences and the inherent difficulty in maintaining long-range temporal coherence without catastrophic forgetting. While existing systems attempt to extend generation duration through segment-wise stitching or sliding-window denoising strategies, these approaches frequently introduce visible artifacts, including temporal drift~\cite{zhang2025frame, gao2025wan}, scene inconsistency, and progressive identity degradation as the video length increases.

Within the specific domain of music-to-dance generation, research has predominantly evolved along two distinct paradigms, each facing significant limitations. The first, known as the \emph{music-to-motion} pipeline, involves predicting 3D skeletal trajectories from audio inputs followed by a separate rendering stage~\cite{tseng2023edge, li2024lodge, li2024lodge++, siyao2022bailando, huang2024beat, luo2024popdg, wang2025dance, wang2025choreomuse, liu2025video}. While this decoupled approach offers explicit control over pose dynamics, it is inherently constrained by the brevity of generated clips (typically under 20 seconds) and often suffers from rendering artifacts and insufficient motion richness. Conversely, the second paradigm employs \emph{end-to-end} methods~\cite{zhang2025let, wang2025dance} that inherit the temporal constraints of general video diffusion systems. These approaches struggle to generate coherent long sequences, frequently exhibiting identity flickering, degraded spatial resolution, and repetitive motion patterns due to an inadequate modeling of long-horizon rhythmic structures.

Addressing this core objective requires establishing a robust mapping between the auditory modality and the visual modality, demanding models that can capture fine-grained cross-modal correlations—aligning rhythm, tempo, and emotional cues with complex spatiotemporal kinematic patterns—while preserving high-fidelity visual details. Although early approaches were hindered by scalability issues and poor synchronization over extended sequences~\cite{tseng2023edge, li2024lodge, li2024lodge++, siyao2022bailando, huang2024beat, luo2024popdg, wang2025dance, wang2025choreomuse, liu2025video, zhang2025let}, recent innovations have begun to overcome these bottlenecks through novel architectural designs. For instance, X-Dancer~\cite{chen2025x} leverages 2D pose tokens within a unified transformer-diffusion framework to enhance expressive synthesis without the errors associated with 3D estimation, while MusicInfuser~\cite{hong2025musicinfuser} pioneers the adaptation of text-to-video backbones with specialized music-video cross-attention modules, marking a critical step toward end-to-end music-to-dance generation. Consequently, the synthesis of \emph{long-duration, high-resolution, and temporally coherent} dance videos that remain strictly synchronized with complex musical compositions remains a significant open challenge.


To address the challenges of temporal discontinuity, low-resolution, and limited duration in existing music-to-dance generation, we propose a hierarchical framework that decouples global motion planning from local refinement. By leveraging global musical context to guide long-term rhythm, our approach effectively eliminates jitter and ensures coherent choreography over extended sequences. By implementing dynamic frame rate adaptation, we generate keyframe sequences that align with various music durations. To improve visual fidelity during rapid movements, we use an optical flow-based loss function combined with motion-speed control, allowing stable minute-scale synthesis at 720p/30fps across five dance genres.

Our specific contributions are summarized as follows:

\begin{itemize}
    \item \textbf{Hierarchical Global-to-Local Generation:} A novel decoupled architecture that uses global musical context for long-term rhythm planning and local refinement for detail, effectively eliminating temporal discontinuities and repetitive artifacts.
    \item \textbf{Dynamic Frame Rate Adaptation:} A mechanism that maps RoPE embeddings to absolute time, enabling the dynamic generation of keyframe sequences adapted to arbitrary music durations for precise temporal alignment.
    \item \textbf{Efficient Choreography Customization:} A LoRA-based fine-tuning strategy that allows users to replicate specific dance routines with only a few example videos, avoiding extensive retraining or large datasets.
\end{itemize}

\section{Related Works}
\label{sec:related_works}

Music-driven dance generation has evolved into a sophisticated domain within multimodal AI, primarily following two paradigms: a two-stage approach generating 3D skeletons for subsequent rendering, and an end-to-end approach synthesizing pixel-level videos directly from audio. The former prioritizes motion controllability, while the latter emphasizes visual fidelity and pipeline simplicity.

\subsection{Two-Stage Paradigm: 3D Pose Generation and Rendering.} This framework decouples motion synthesis from appearance rendering, enabling precise alignment between musical features and kinematic dynamics.

\subsubsection{3D Skeleton Sequence Generation.} Research initially focused on mapping audio to 3D joint rotations using Transformers or VAEs. For solo dance, seminal works include EDGE~\cite{tseng2023edge} and LODGE~\cite{li2024lodge}, which leverage motion priors and hierarchical latents for diversity and coherence. Bailando~\cite{siyao2022bailando} employs reinforcement learning to ensure physical plausibility, while Beat-It~\cite{huang2024beat} and POPDG~\cite{luo2024popdg} refine beat synchronization and stylistic consistency. Extending to group scenarios, AIOZ-GDANCE~\cite{le2023music} models inter-dancer spatial interactions, and Duolando~\cite{siyao2024duolando} utilizes dual-stream architectures to coordinate individual dynamics with collective formations.

\subsubsection{Integrated Rendering Pipelines.} Recent studies integrate motion synthesis with neural renderers to produce complete videos. Video Motion Graphs~\cite{liu2025video} pioneered clip retrieval for smooth transitions. X-Dancer~\cite{chen2025x} combines high-fidelity motion generation with photorealistic neural rendering, while ChoreoMuse~\cite{wang2025choreomuse} introduces style transfer capabilities during rendering.

\subsection{End-to-End Paradigm: Direct Video Generation.} Leveraging large-scale video diffusion models, this paradigm generates dance videos directly from music, bypassing explicit 3D intermediates to enhance textural realism.

\subsubsection{Foundational Video Generation Technologies.} Direct synthesis relies on advancements in general and audio-driven video generation. Models like Wan~\cite{wan2025wan}, Hunyuan~\cite{wu2025hunyuanvideo}, and Seedance~\cite{seedance2025seedance} demonstrate strong capabilities in coherent, high-resolution synthesis from text or dance-specific data. KlingAvatar 2.0~\cite{team2025klingavatar} and Wan-S2V~\cite{gao2025wan} excel in maintaining character identity. To address long-term consistency, FramePack~\cite{zhang2025frame} and TempoMaster~\cite{ma2025tempomaster} introduce advanced temporal attention mechanisms. Furthermore, OmniHuman1.5~\cite{jiang2025omnihuman}, VideoJAM~\cite{chefer2025videojam}, and STG~\cite{hyung2025spatiotemporal} focus on anatomical correctness and spatiotemporal optimization to reduce flickering and enhance structural integrity.

\subsubsection{Direct Music-to-Dance Methods.} Building on these foundations, specialized methods directly map music to video. MusicInfuser~\cite{hong2025musicinfuser} integrates audio embeddings deeply into the diffusion process for intrinsic rhythm alignment. MVAA~\cite{zhang2025let} utilizes multi-view audio attention to synchronize complex rhythmic patterns with camera and dancer movements. Dance Any Beat~\cite{wang2025dance} emphasizes zero-shot generalization across unseen genres. While these approaches offer streamlined pipelines ideal for creative applications, they currently face challenges in long-term and high-resolution dance video generation.

\section{Method}
\label{sec:method}

\begin{figure}[tb]
  \centering
  \includegraphics[width=\linewidth]{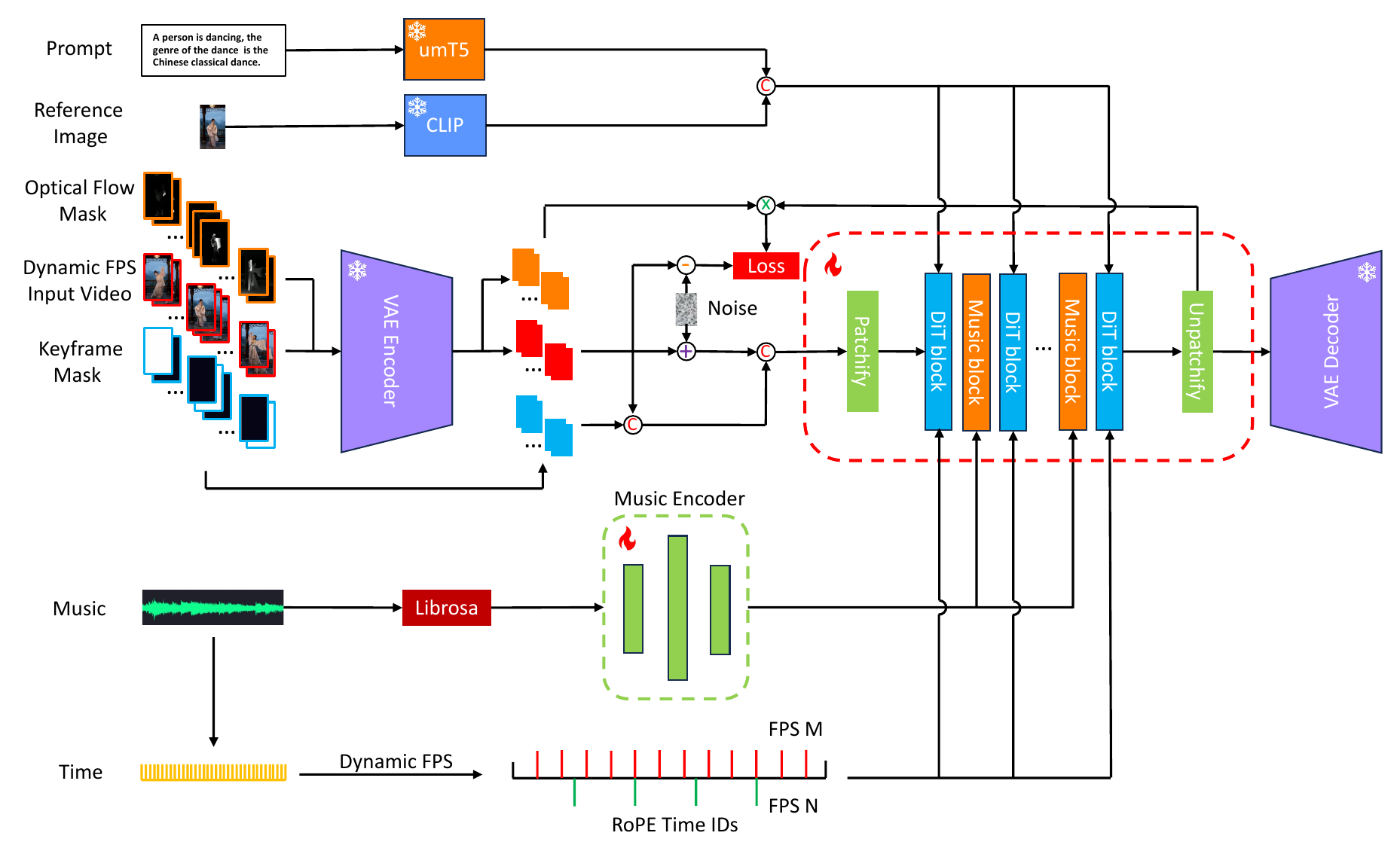}
  \caption{Training Pipeline. Our training strategy focuses on optimizing the Diffusion Transformer (DiT) blocks and the Music Encoder. Both the global keyframe generation and local temporal refinement stages adhere to a unified training pipeline. The primary distinction lies in the input video, keyframe mask, prompt, and the target frame rates (fps) specific to each stage's temporal granularity.
  }
  \label{fig:pipeline_train}
  \vspace{-10pt}
\end{figure}

\subsection{Model Architecture}

Our proposed framework, depicted in Fig.~\ref{fig:pipeline_train}, is built upon the foundational architecture of Wan-I2V~\cite{wan2025wan}, which serves as the backbone for our music-to-dance generation pipeline. 

Our model employs a unified training framework that seamlessly integrates both global and local learning objectives within the same network. To achieve this unification, we construct the input tensor by concatenating a keyframe mask with the video latents encoded by the VAE encoder~\cite{wan2025wan}. This mask serves as a dynamic control signal to distinguish between the two training phases:

\paragraph{Global Phase.} To simulate long-term dependency modeling, the keyframe mask is initialized with all zeros except for the first frame, which is set to one. This configuration forces the model to infer the entire sequence's structural evolution based solely on the initial condition.

\paragraph{Local Phase.} To train the model on fine-grained interpolation and short-term dynamics, the keyframe mask indices are randomly selected from the sequence and set to one, while all other positions remain zero. This stochastic masking strategy compels the network to learn robust motion synthesis conditioned on sparse, arbitrary temporal anchors rather than just the starting frame.\\

This flexible masking mechanism allows the union training pipeline to effectively learn both holistic choreographic planning and detailed frame-to-frame continuity. 

To enhance temporal adaptability, we implement a dynamic frame rate injection mechanism inspired by recent advancements in large-scale vision-language models~\cite{bai2025qwen25vltechnicalreport}. Specifically, we integrate absolute time identifiers directly into the Rotary Positional Embedding (RoPE)~\cite{heo2024rotary} module, enabling the model to precisely contextualize frames regardless of variable sampling rates. 

Regarding multimodal conditioning, we employ a dedicated lightweight music encoder~\cite{li2024lodge++} to extract latent acoustic features from the input music. These encoded musical representations are then seamlessly injected into the Diffusion Transformer (DiT) blocks by the Music block~\cite{gao2025wan}, facilitating robust audio-visual alignment and ensuring that the generated motion dynamics remain strictly synchronized with the rhythmic and structural nuances of the accompanying soundtrack.

\subsection{Training Objective}

Our framework integrates three distinct modalities: music, text prompts, and keyframes. Building upon the training pipeline established by Wan-I2V~\cite{wan2025wan}, we adopt a Rectified Flow (RF) formulation~\cite{esser2024scaling} to model the generative process. 

Given video latents $x_1$, Gaussian noise $x_0 \sim \mathcal{N}(0, I)$, and a timestep $t \in [0,1]$ sampled from a logit-normal distribution, we construct an intermediate latent $x_t$ via linear interpolation:

\begin{equation}
  x_t = t x_1 + (1 - t) x_0.
  \label{eq:xt}
\end{equation}

The corresponding ground truth velocity field $v_t$ is defined as the derivative of the trajectory:

\begin{equation}
  v_t = \frac{dx_t}{dt} = x_1 - x_0.
  \label{eq:vt}
\end{equation}

The model is optimized to predict this velocity vector. Consequently, our objective function is formulated as the Mean Squared Error (MSE) between the predicted velocity and the ground truth:

\begin{equation}
  \mathcal{L} = \mathbb{E}_{x_0, x_1, c_{ref}, c_{txt}, f_m, t} \| (u(x_t, c_{ref}, c_{txt}, f_m, t; \theta) \odot w_{optical\_flow}  - v_t)\|^2,
  \label{eq:Loss}
\end{equation}

where:
\begin{itemize}
    \item $c_{ref}$ denotes the visual conditioning features extracted from the reference image using CLIP~\cite{radford2021learning}.
    \item $c_{txt}$ represents the textual embedding sequence (512 tokens) encoded by umT5~\cite{chung2023unimax}.
    \item $f_m$ signifies the acoustic features derived from the input music via our training music encoder.
    \item $\theta$ corresponds to the learnable parameters of the network;
    \item $u(\cdot)$ is the model's predicted velocity output conditioned on the latent state, multimodal inputs, and timestep.
    \item $\odot$ denotes the element-wise multiplication.
    \item $w_{optical\_flow}$ represents the optical flow latents encoded by the VAE~\cite{wan2025wan} encoder corresponding to the input video.
\end{itemize}

This formulation ensures that the model effectively learns to map the joint multimodal context to a coherent temporal evolution of video latents.

\section{Experiments}
\label{sec:experiments}

\subsection{Implementation Details}

\subsubsection{Dataset.}
To facilitate high-resolution dance video generation, we curated a proprietary dataset of approximately 200 hours of high-quality video (minimum resolution of 720p at 30 fps), intentionally excluding datasets like AIST~\cite{tsuchida2019aist} and Finedance~\cite{li2023finedance} due to their limitations in duration, resolution, and lack of raw high-definition video for synthesis tasks. Our dataset includes five dance genres—Chinese Classical, K-Pop, Latin, Tap, and Street dance—ensuring a nearly uniform temporal distribution to mitigate class imbalance.

For training, we employed a data segmentation strategy, dividing raw videos into 5-second clips with a 50\% overlap, enhancing short-term motion dynamics analysis. We also processed audio tracks using the Librosa library~\cite{mcfee2015librosa} for efficient feature extraction. To improve motion fidelity, we generated optical flow masks with SEA-RAFT~\cite{wang2024sea}, integrating them into the loss function to guide accurate motion synthesis.

Additionally, we stratified the dataset based on kinematic velocity derived from the keypoints~\cite{xu2022vitpose}, categorizing motions into slow, medium, and high tiers (10\% slow, 80\% medium, 10\% high) to encourage learning of realistic movement patterns while focusing on medium-speed motions, characteristic of fluid human dance.

\subsubsection{Model Training Details.}
We implement a two-stage training protocol to enhance computational efficiency and output fidelity.

\paragraph{Low-Resolution Pre-training.} In the initial phase, we pre-train the global and local components using videos resized to $320 \times 544$ pixels based on Wan-I2V~\cite{wan2025wan}, fitting within a single NVIDIA A100 (80GB) GPU's memory constraints. This low-resolution model is trained on 128 A100 GPUs for 20,000 steps at a learning rate of $1 \times 10^{-5}$. The focus is on aligning human dance kinematics with musical rhythms before tackling high-frequency details, with optical flow loss applied only during local model training.

\paragraph{High-Resolution Fine-tuning.} Next, we fine-tune the pretrained weights using native 720p video data. To manage the memory overhead, we utilize USP~\cite{fang2024usp} with a Ulysses degree and sequence parallelism degree of 8. This stage runs on 128 A100 GPUs for 4,000 steps, maintaining a learning rate of $1 \times 10^{-5}$. The result is a generative framework capable of producing high-fidelity, synchronized 720p dance videos.

\paragraph{Customize dancing via LoRA.} We also introduce a lightweight adaptation mechanism for mimicking different dance styles based on 16 reference videos with identical choreography. Using Low-Rank Adaptation (LoRA)~\cite{hu2022lora} with a rank of 32, we restrict training to 800 steps at a learning rate of $1 \times 10^{-4}$ to develop a specialized mimicry model quickly without losing the base model's capabilities.

\subsubsection{Model Inference Details.}

\begin{figure}[tb]
  \centering
  \includegraphics[width=\linewidth]{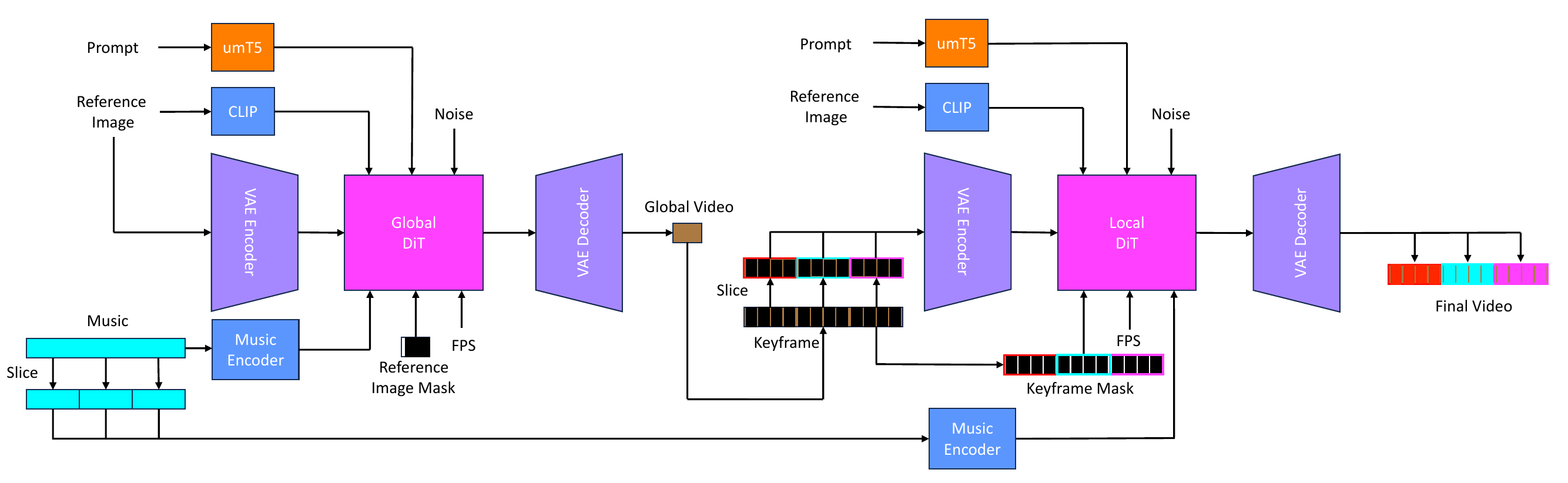}
  \caption{Inference Pipeline. Initially, the global stage synthesizes a sparse keyframe global video that captures the overall motion structure and rhythmic alignment. Subsequently, this global output serves as a structural guidance condition for the local stage, which interpolates and refines the sequence to generate the final high-frame-rate, dense video with enhanced temporal coherence and visual detail.
  }
  \label{fig:pipeline_infer}
  \vspace{-10pt}
\end{figure}

As illustrated in Fig.~\ref{fig:pipeline_infer}, our inference pipeline operates through a hierarchical Global-to-Local strategy.

\paragraph{Global Keyframe Generation.} The input music, text prompt, random noise, dynamic FPS, reference image, and mask are encoded into the Global Diffusion Transformer (DiT). After 50 denoising iterations, we generate a sparse global keyframe video comprising 38 frames, capturing the overall choreographic structure and temporal coherence.

\paragraph{Temporal Segmentation and Preparation.} The 38 keyframes are evenly distributed across the video duration and segment the full sequence into clips of 149 frames centered around each keyframe. Simultaneously, the audio is sliced into corresponding 5-second segments.

\paragraph{Local Frame Generation.} Each segment's sliced music, keyframes, keyframe masks, text prompt, reference mask, and new noise are input into the Local DiT. After another 50 iterations, high-fidelity intermediate frames are produced, preserving the global keyframes, with this process running in parallel across segments for efficiency.

\paragraph{Final Assembly.} All generated local video clips are concatenated in order, resulting in a coherent dance video that balances detailed motion with global structural consistency.

\subsection{Experimental Results}

\subsubsection{Minute-scale Music-to-Dance Generation.} 

\begin{figure}[tb]
  \centering
  \includegraphics[width=\linewidth]{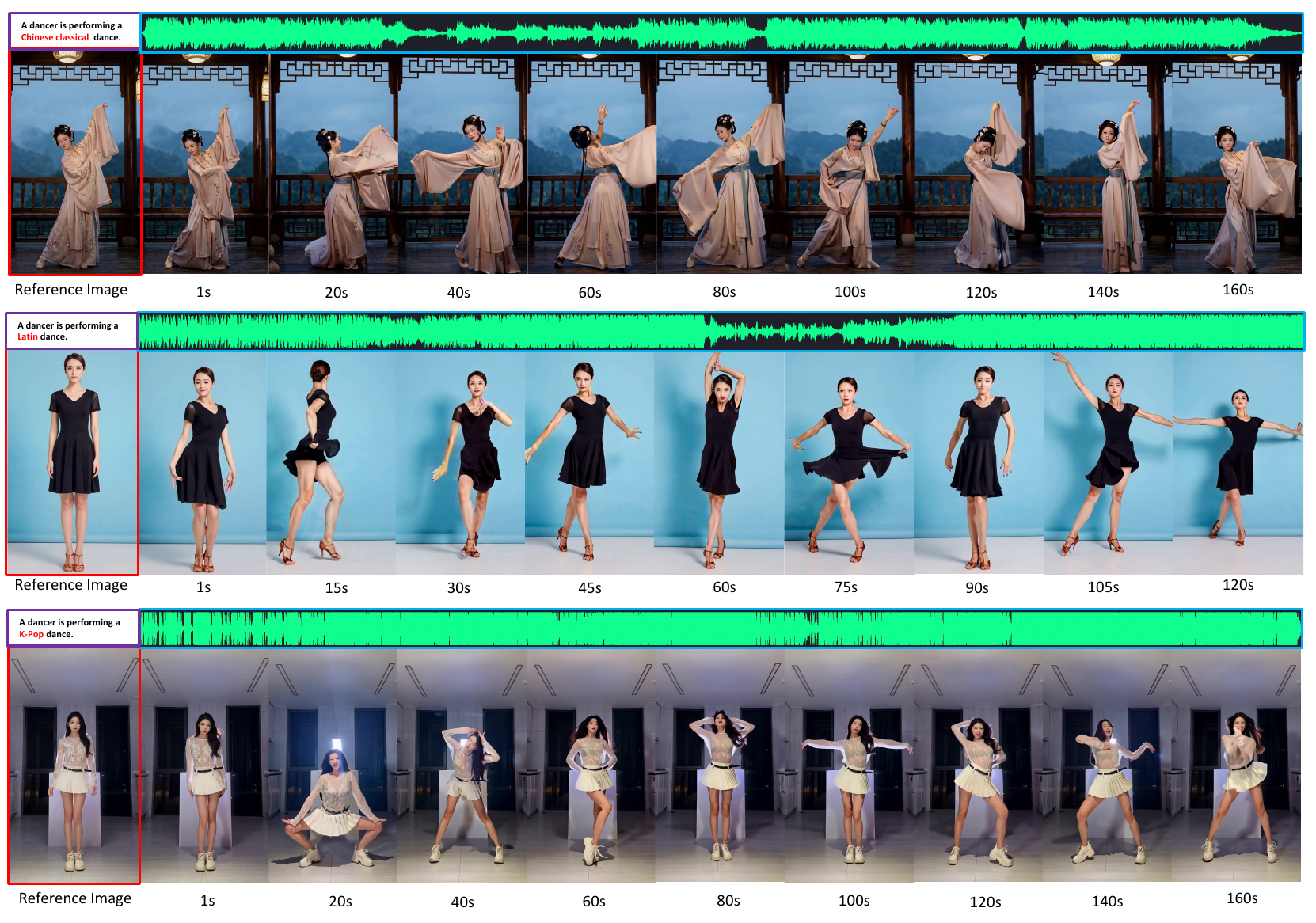}
  \caption{Minute-scale Video Generation. Our framework enables the synthesis of dance videos exceeding one minute in duration across diverse genres. The first row features Chinese Classical dance with fluid movements, the second showcases the energetic rhythms of Latin dance, and the third reflects the unique elements of K-Pop dance. 
  }
  \label{fig:figure_long_video}
  \vspace{-10pt}
\end{figure}

Our framework leverages a global-to-local architecture to synthesize minute-scale dance videos based on input music. As shown in Fig.~\ref{fig:figure_long_video}, it maintains high fidelity and spatial resolution for sequences up to 160 seconds while preserving identity across durations; generated subjects closely match the reference images in fine details like clothing textures.

Additionally, the model adheres precisely to semantic prompts for dance genres. The visual results demonstrate accurate representation of choreographic styles, which highlights the model’s ability to align high-level textual instructions with specific motion patterns. 

By leveraging a hierarchical global-to-local architecture, we achieve high-definition output while effectively mitigating the error accumulation and temporal drift, and maintaining the dance style.

\subsubsection{Customized Dance Generation.}
\begin{figure}[tb]
  \centering
  \includegraphics[width=\linewidth]{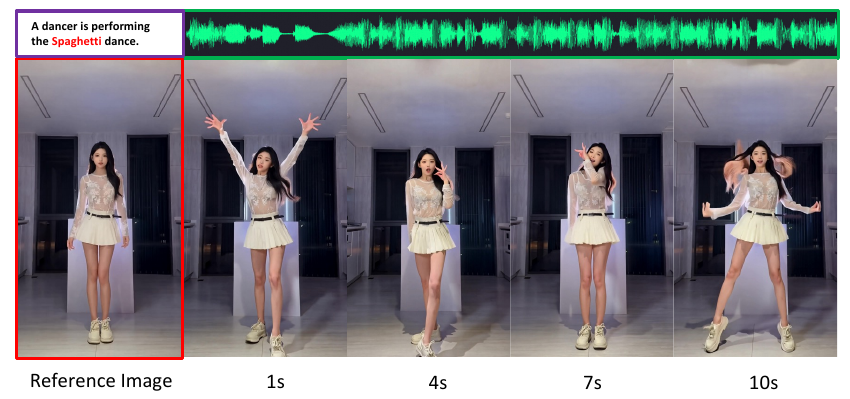}
  \caption{Customized Choreography. Leveraging Low-Rank Adaptation (LoRA), our framework enables the efficient fine-tuning of generative models to mimic specific human choreographies.}
  \label{fig:figure_lora}
\end{figure}

Our framework not only synthesizes dance videos based on input music but also generates specific choreographic motions using Low-Rank Adaptation (LoRA)~\cite{hu2022lora}. As shown in Fig.~\ref{fig:figure_lora}, this method aligns generated sequences with human-curated motion templates. For instance, it successfully reproduces the unique kinematic characteristics of the "Spaghetti" dance, demonstrating LoRA's effectiveness in fine-tuning motion control.

\subsection{Evaluations and Comparisons}

\subsubsection{Quantitative Evaluation.}
In the music-to-dance generation field, much of the literature, including EDGE~\cite{tseng2023edge}, LODGE++\cite{li2024lodge++}, focuses on synthesizing intermediate 3D representations rather than final video outputs. Therefore, our comparison is limited to end-to-end frameworks that directly synthesize videos, specifically X-Dancer~\cite{chen2025x} and MusicInfuser~\cite{hong2025musicinfuser}. 

We followed the evaluation protocol from MusicInfuser~\cite{hong2025musicinfuser}, which measures performance in three key areas: Dance Quality, Video Quality, and Prompt Alignment.

\paragraph{Dance Quality.} As detailed in Table~\ref{tab:dance_quality_metrics}, our method demonstrates superior performance over prior art across all key metrics, including style alignment, beat synchronization, body representation fidelity, movement realism, and choreography complexity.
\paragraph{Video Quality.} In terms of visual fidelity, our approach consistently outperforms baseline methods, achieving higher scores in imaging quality, aesthetic appeal, and temporal consistency. See more details in Table~\ref{tab:video_quality_metrics}.
\paragraph{Prompt Alignment.} Since X-Dancer~\cite{chen2025x} lacks native support for textual prompt conditioning, this specific metric was evaluated solely against MusicInfuser~\cite{hong2025musicinfuser}. Our results indicate a marked improvement in style capture, creative interpretation, and overall user satisfaction, underscoring the enhanced controllability of our framework. The details are illustrated in Table~\ref{tab:prompt_alignment_metrics}.

\begin{table}[tb]
  \centering
  \begin{minipage}{\textwidth}
  \caption{Dance quality metrics comparing different models. A, I, and T denote audio, image, and text input modalities, respectively.
  }
  \label{tab:dance_quality_metrics}
  \centering
  \resizebox{\columnwidth}{!}{
  \begin{tabular}{@{}cccccccc@{}}
    \toprule
    \multirow{2}{*}{\textbf{Model}} & \multirow{2}{*}{\textbf{Modality}} & \textbf{Style} & \textbf{Beat} & \textbf{Body} & \textbf{Movement} & \textbf{Choreography} & \multirow{2}{*}{\textbf{Average}} \\
    & & \textbf{Alignment} & \textbf{Alignment} & \textbf{Representation} & \textbf{Realism} & \textbf{Complexity}\\
    \midrule
    MusicInfuser~\cite{hong2025musicinfuser} & A+T & 7.43 & 7.26 & 4.76 & 6.89 & 4.80 & 6.23 \\
    X-Dancer~\cite{chen2025x} & A+I & 6.55 & 6.78 & 6.23 & 6.11 & 4.63 & 6.06 \\
    Wan-Dancer(Ours) & A+T+I & \textbf{8.33} & \textbf{8.73} & \textbf{9.01} & \textbf{9.23} & \textbf{6.98} & \textbf{8.46} \\
    \bottomrule
  \end{tabular}
  }
  \end{minipage}
  \begin{minipage}{0.48\textwidth}
  \centering
  \caption{Video quality metrics comparing different models.
  }
  \resizebox{\columnwidth}{!}{
  \label{tab:video_quality_metrics}
  \begin{tabular}{@{}cccccc@{}}
    \toprule
    \multirow{2}{*}{\textbf{Model}} & \multirow{2}{*}{\textbf{Modality}} & \textbf{Imaging} & \textbf{Aesthetic} & \textbf{Overall} & \multirow{2}{*}{\textbf{Average}} \\
    & & \textbf{Quality} & \textbf{Quality} & \textbf{Consistency} \\
    \midrule
    MusicInfuser~\cite{hong2025musicinfuser} & A+T & 5.03 & 4.74 & 5.88 & 5.22 \\
    X-Dancer~\cite{chen2025x} & A+I & 5.23 & 6.05 & 7.48 & 6.23 \\
    Wan-Dancer(Ours) & A+T+I & \textbf{6.84} & \textbf{7.91} & \textbf{7.63} & \textbf{7.46} \\
    \bottomrule
  \end{tabular}
  }
  \end{minipage}%
  \begin{minipage}{0.48\textwidth}
  \centering
  \caption{Prompt alignment metrics comparing different models.
  }
  \resizebox{\columnwidth}{!}{
  \label{tab:prompt_alignment_metrics}
  \centering
  \begin{tabular}{@{}ccccc@{}}
    \toprule
    \multirow{2}{*}{\textbf{Model}} & \textbf{Style} & \textbf{Creative} & \textbf{Overall} & \multirow{2}{*}{\textbf{Average}} \\
    & \textbf{Capture} & \textbf{Interpretation} & \textbf{Satisfaction} \\
    \midrule
    MusicInfuser~\cite{hong2025musicinfuser} & 5.03 & 7.48 & 7.33 & 6.61 \\
    Wan-Dancer(Ours) & \textbf{9.14} & \textbf{8.73} & \textbf{9.21} & \textbf{9.03} \\
    \bottomrule
  \end{tabular}
  }
  \end{minipage}%
\end{table}

\subsubsection{Qualitative Comparisons.}

\begin{figure}[tb]
  \centering
  \begin{subfigure}{\linewidth}
    \includegraphics[width=\linewidth]{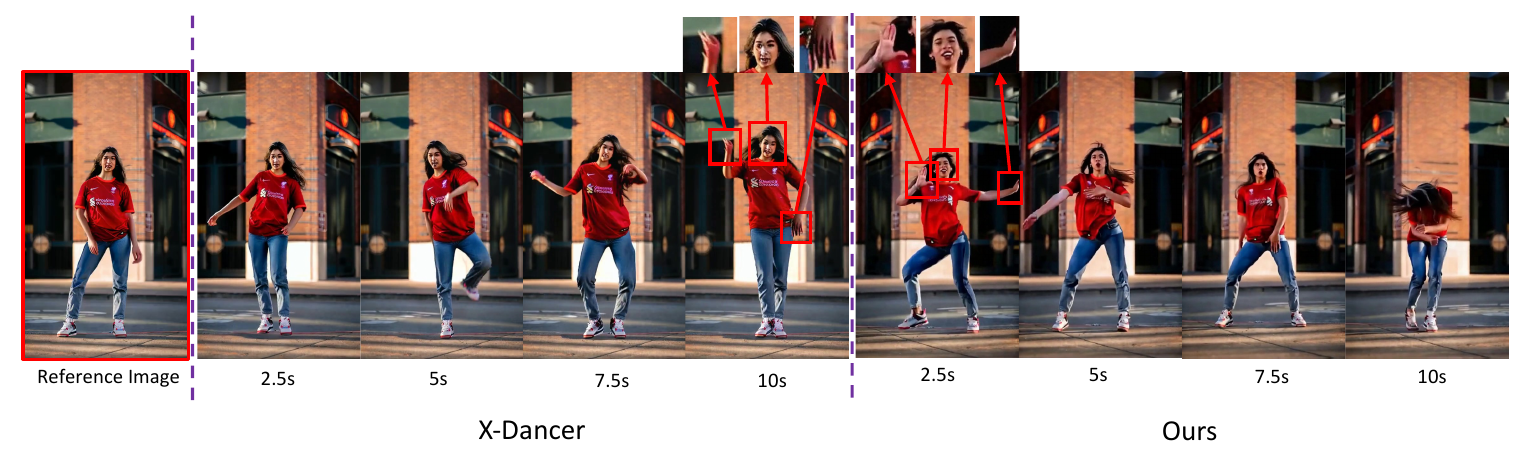}
    \caption{Comparison with X-Dancer~\cite{chen2025x}. Our approach yields markedly sharper reconstructions of complex anatomical regions, particularly hands and facial features, which often suffer from blurring or distortion in X-Dancer.}
    \label{fig:figure_compare_xdancer}
  \end{subfigure}
  \hfill
  \begin{subfigure}{\linewidth}
    \includegraphics[width=\linewidth]{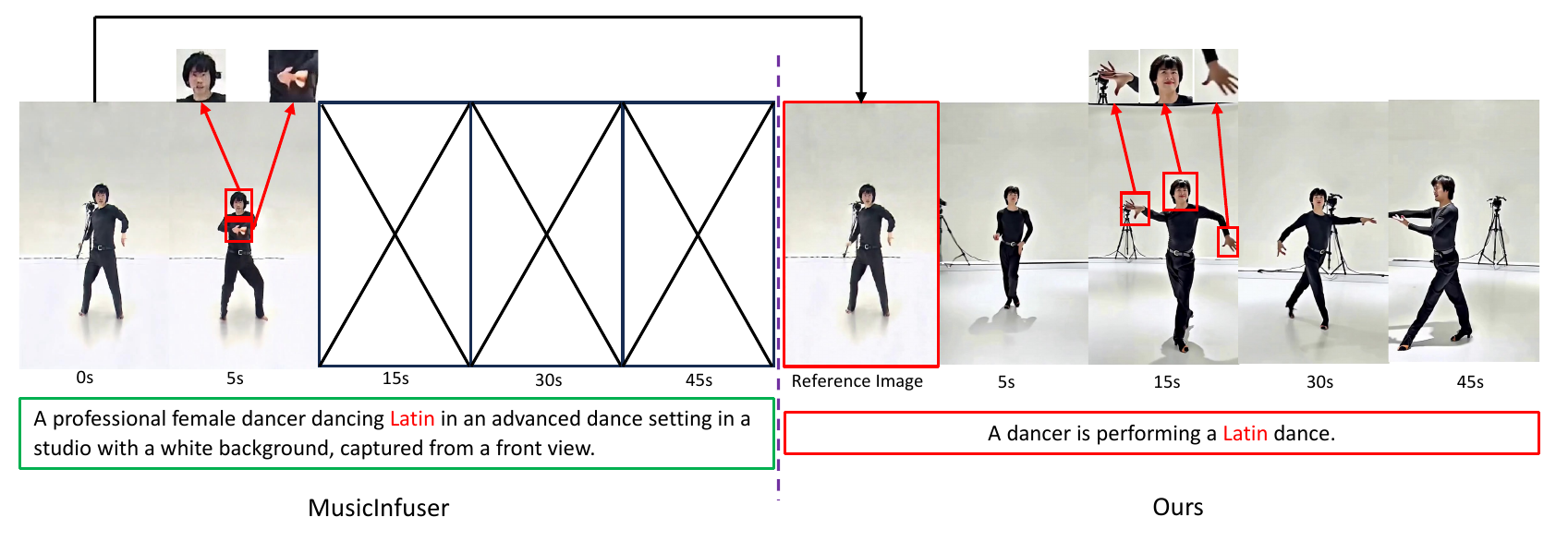}
    \caption{Comparison with MusicInfuser~\cite{hong2025musicinfuser}. Unlike MusicInfuser, which is limited to short temporal windows, our framework generates minute-long dance videos with sustained coherence. It also delivers higher spatial resolution, with sharper facial and hand details where baselines often blur, and achieves stronger semantic alignment, ensuring motions strictly follow the text prompts while staying rhythmically synchronized with the audio.}
    \label{fig:figure_compare_musicinfuser}
  \end{subfigure}
  \hfill
  \vspace{-10pt}
  \label{fig:figure_compare_with_sota}
  \caption{Comparison with X-Dancer~\cite{chen2025x} and MusicInfuser~\cite{hong2025musicinfuser}.}
  \vspace{-10pt}
\end{figure}

We conducted a comprehensive qualitative evaluation comparing our framework against two SOTA baselines: X-Dancer~\cite{chen2025x} and MusicInfuser~\cite{hong2025musicinfuser}.

In comparison with X-Dancer~\cite{chen2025x} (Fig.~\ref{fig:figure_compare_xdancer}), our method excels in visual fidelity, controllability, and temporal coherence. X-Dancer often shows structural artifacts in facial and hand features, while our approach maintains anatomical integrity. Without textual conditioning, X-Dancer's style generation is limited, whereas our framework effectively synthesizes five dance genres based on specific prompts. Additionally, X-Dancer relies on short, disjointed clips, causing noticeable temporal discontinuities, while our model utilizes global music representations for seamless long-term coherence.

When compared to MusicInfuser~\cite{hong2025musicinfuser} (Fig.~\ref{fig:figure_compare_musicinfuser}), our method stands out in temporal scalability and spatial resolution. MusicInfuser can only generate short 5-second segments and often provides blurred or indistinct details. In contrast, our method captures intricate details with exceptional clarity and aligns movements accurately with specified dance styles, while MusicInfuser struggles with stylistic consistency, resulting in generic outputs.

\subsubsection{Ablation Studies.}

\begin{figure}[tb]
  \centering
  \begin{subfigure}{0.48\linewidth}
    \includegraphics[width=\linewidth]{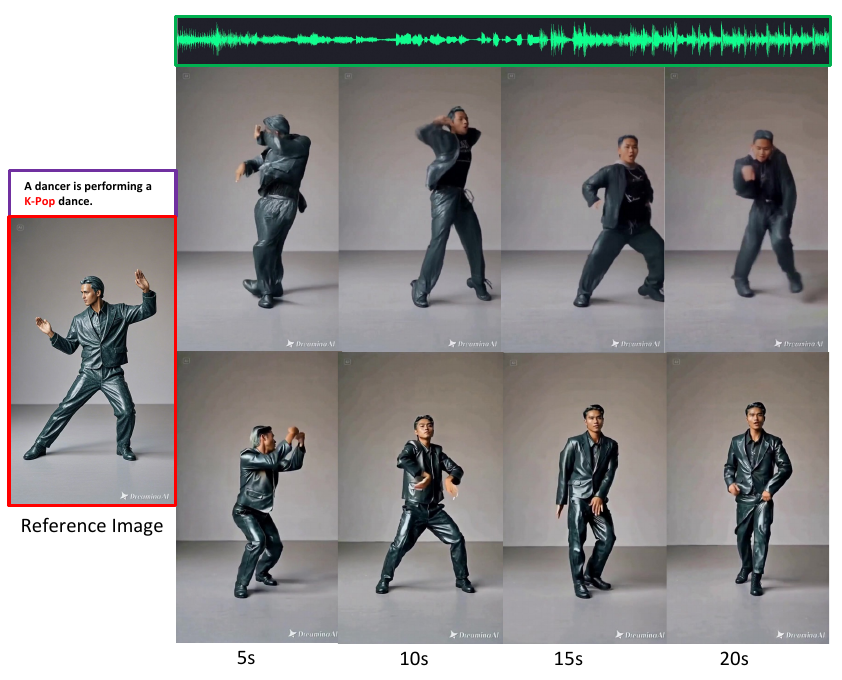}
    \caption{Long-Duration Comparison. The top row displays a naive baseline with 5-second segments generated sequentially, leading to discontinuities and errors. In contrast, the bottom row shows our hierarchical method, which improves temporal stability and consistency, eliminating drift and artifacts.}
    \label{fig:figure_compare_long_video}
  \end{subfigure}
  \begin{subfigure}{0.48\linewidth}
    \includegraphics[width=\linewidth]{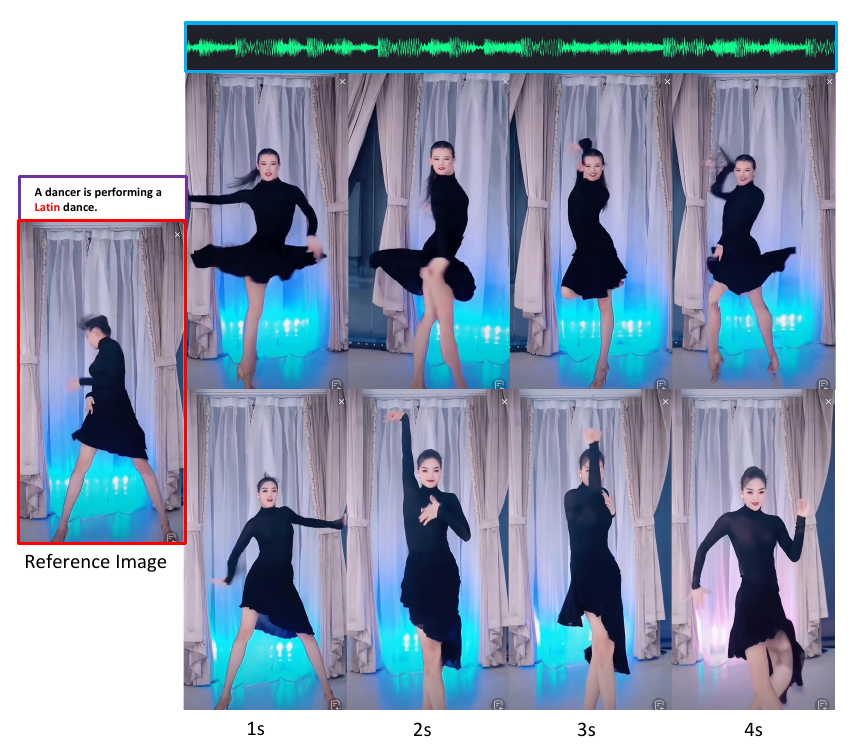}
    \caption{Ablation on Optical Flow Loss Weight: The top row displays results from a model without optical flow loss weight, leading to motion blur and detail loss during rapid movements. In contrast, the bottom row from our model with optical flow loss shows sharper textures and better structure, particularly in dynamic areas like the hands, minimizing smearing artifacts and enhancing clarity.}
    \label{fig:figure_compare_optical_flow_loss_weight}
  \end{subfigure}
  \label{fig:figure_ablation_study_01}
  \caption{Comparisons for long-duration comparison and optical flow loss weight.}
  \vspace{-10pt}
\end{figure}

To rigorously evaluate the efficacy of our proposed architecture, we conducted a comprehensive ablation study isolating the contributions of four critical components: long-term temporal coherence mechanisms, optical flow loss weighting, dynamic frame rate injection, and motion speed stratification.

\paragraph{Long-Term Temporal Coherence.} Our global-to-local strategy enhances minute-scale video synthesis by using holistic musical context to guide temporal evolution. Relying solely on local processing fails to maintain choreographic continuity over longer durations, leading to disjointed segments and significant semantic drift due to error accumulation. As shown in Fig.~\ref{fig:figure_compare_long_video}, naive clip concatenation leads to drastic deviations in subject appearance. In contrast, our framework demonstrates that strong long-term modeling is essential for coherent extended dance performances, preserving motion fluidity and consistent identity.

\paragraph{Optical Flow Loss Weighting.} Integrating optical flow constraints from SEA-RAFT~\cite{wang2024sea} greatly improves spatial consistency and edge fidelity. Fig.~\ref{fig:figure_compare_optical_flow_loss_weight} shows that excluding this weighted loss results in blurred boundaries and structural distortions during rapid motion. These findings emphasize the necessity of motion guidance in the objective function to retain anatomical integrity during dynamic dance maneuvers.

\begin{figure}[tb]
  \centering
  \begin{subfigure}{0.48\linewidth}
    \includegraphics[width=\linewidth]{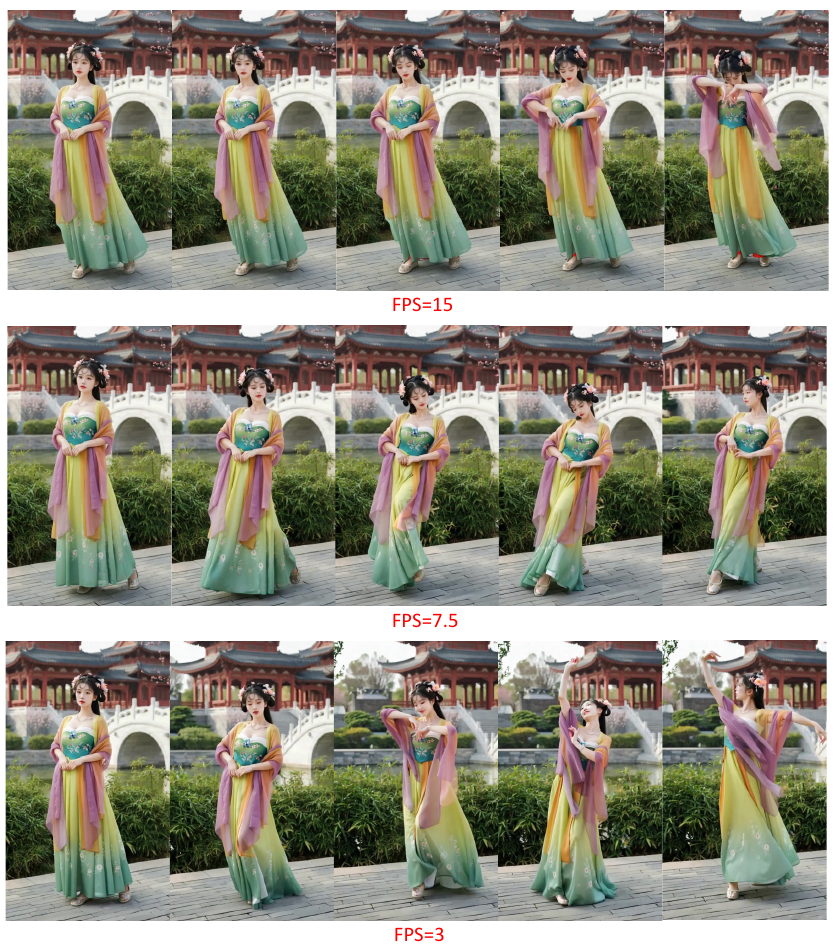}
    \caption{Dynamic Frame Rate Adaptation: The rows show generation results at 15, 7.5, and 3 fps. As the frame rate decreases, temporal sampling becomes sparser, leading to greater kinematic disparities between frames.}
    \label{fig:figure_compare_dynamic_fps}
  \end{subfigure}
  \begin{subfigure}{0.48\linewidth}
    \includegraphics[width=\linewidth]{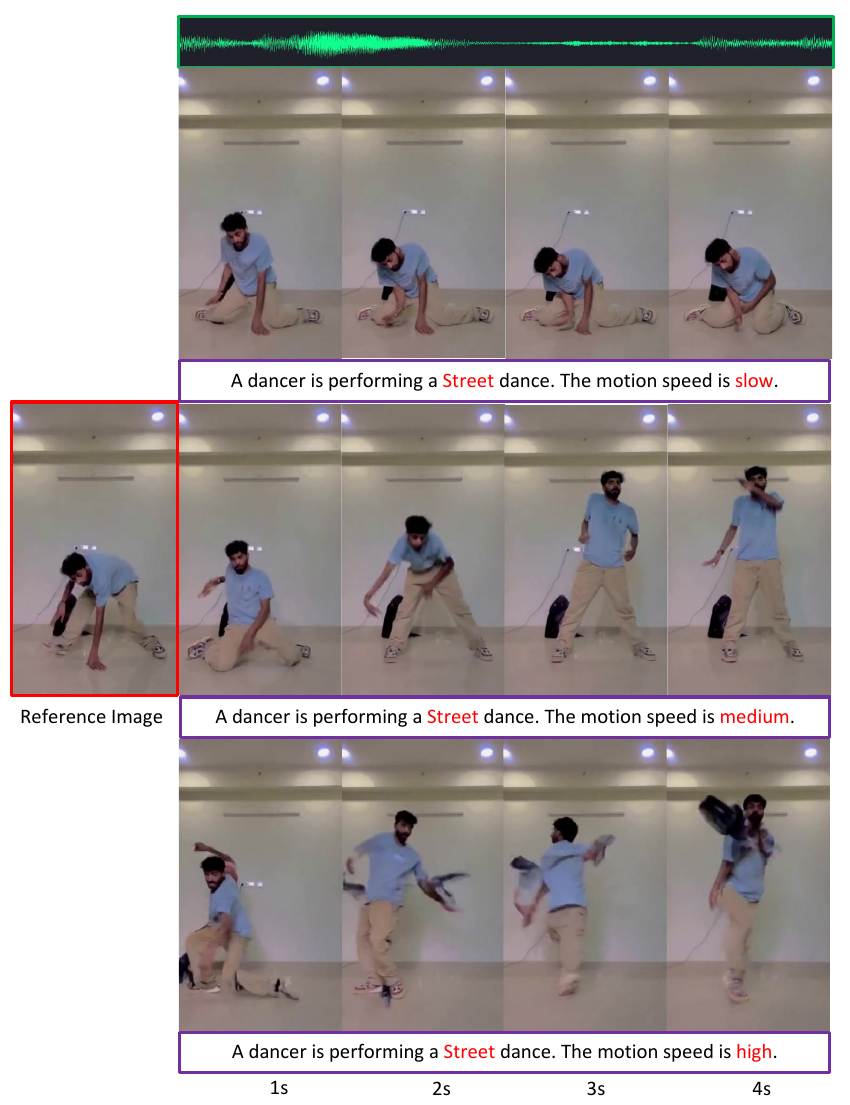}
    \caption{The rows illustrate generation results at slow, medium, and high motion speeds. The slow setting misaligns with the musical tempo, while the high-speed setting creates visual artifacts due to complexity. The medium speed strikes an optimal balance.}
    \label{fig:figure_motion_speed}
  \end{subfigure}
  \label{fig:figure_ablation_study_02}
  \caption{Comparisons for dynamic frame rate adaptation and motion speed.}
  \vspace{-10pt}
\end{figure}

\paragraph{Dynamic Frame Rate Injection.} This module adjusts the frame rate according to the music sequence's total duration, integrating this into the Rotary Positional Embedding (RoPE) mechanism~\cite{heo2024rotary}. As illustrated in Fig.~\ref{fig:figure_compare_dynamic_fps}, this adaptive strategy modulates the temporal density between frames, allowing for a range from high (15 fps) to low (3 fps) sampling. This flexibility is vital for generating an optimal number of frames during Local Keyframe Generation, aligning temporal resolution with musical context to maintain fluidity and consistency.

\paragraph{Motion Speed Stratification.} By categorizing training samples into slow, medium, and high-velocity tiers, we prioritized learning naturalistic medium-speed dynamics. Our framework allows precise control over motion dynamics during inference through textual prompts. Fig.~\ref{fig:figure_motion_speed} shows that medium-speed generation strikes the best balance between choreographic rhythm and visual fidelity, while slow or fast sequences are prone to artifacts and inconsistencies. This skewed training data distribution supports superior model convergence and robustness for naturalistic dance movements.

Collectively, these results validate that each component plays a distinct and synergistic role in achieving high-fidelity, temporally consistent, and stylistically accurate dance video generation.

\subsubsection{Diversity.}

\begin{figure}[tb]
  \centering
  \begin{subfigure}{0.32\linewidth}
    \includegraphics[width=\linewidth]{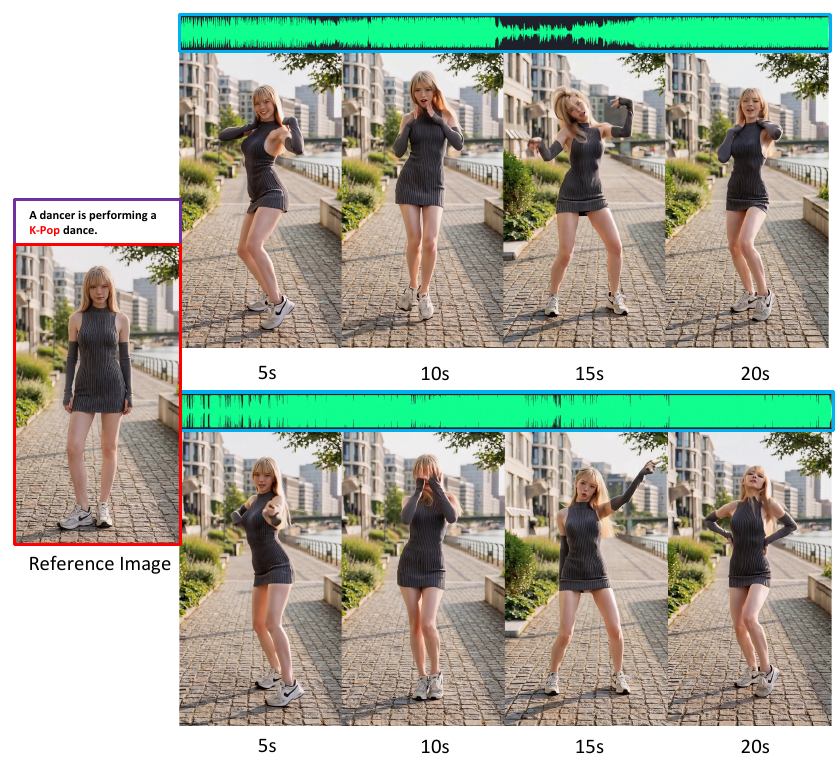}
    \caption{The same reference image with different music. The top and bottom videos use different pieces of music.}
    \label{fig:figure_compare_same_ref_differ_song}
  \end{subfigure}
  \hfill
  \begin{subfigure}{0.33\linewidth}
    \includegraphics[width=\linewidth]{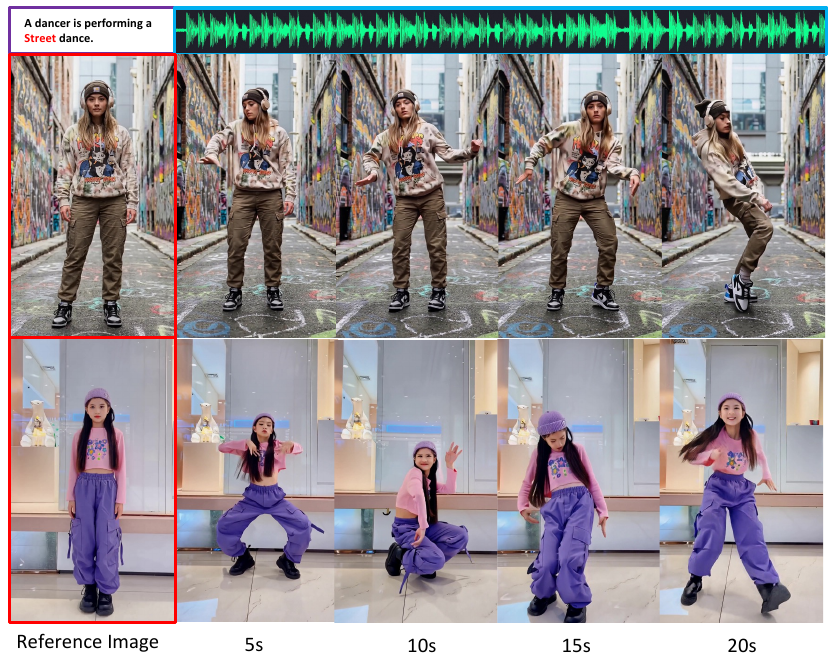}
    \caption{The same music with different reference images. The top and bottom videos use different images.}
    \label{fig:figure_compare_same_song_differ_ref}
  \end{subfigure}
  \hfill
  \begin{subfigure}{0.33\linewidth}
    \includegraphics[width=\linewidth]{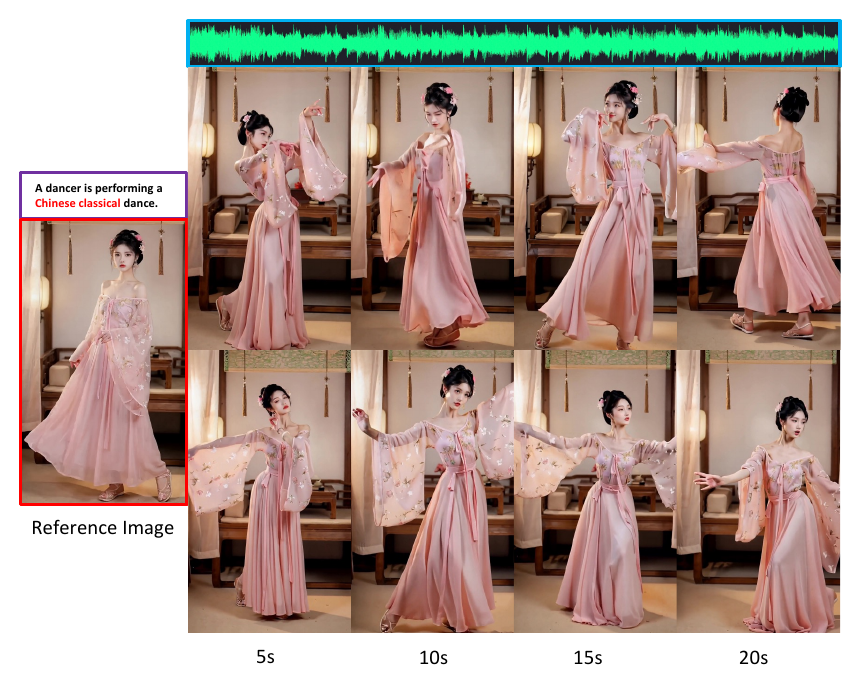}
    \caption{The same music and reference image are used, but different seeds generate videos in the top and bottom rows.}
    \label{fig:figure_compare_same_song_same_ref_diff_seed}
  \end{subfigure}
  \caption{We assess the generative diversity of our method through controlled experiments by varying the reference image, music input, and random seed..}
  \label{fig:figure_diversity}
  \vspace{-10pt}
\end{figure}

To evaluate the generative diversity of our framework, we conducted an empirical analysis in three phases: (1) fixing a reference image while varying the input music; (2) maintaining the same music while changing the reference image; and (3) keeping both constant while varying the noise seed. As shown in Fig.~\ref{fig:figure_diversity}, our model effectively synthesizes coherent dance videos with distinct motion patterns across all scenarios. These results highlight the framework's ability to decouple and recombine multimodal inputs, achieving high variability in choreographic output while maintaining temporal consistency and visual fidelity.

\section{Conclusion}
\label{sec:conclusion}

In conclusion, we have tackled the challenge of generating long-duration, high-definition, rhythmically synchronized dance videos by overcoming limitations in current diffusion models. Our novel hierarchical framework decouples global keyframe planning from local temporal refinement, eliminating issues like temporal drift and identity degradation that impact existing 3D-based and end-to-end approaches. Key innovations, such as dynamic frame rate adaptation, an optical flow-based loss function, and precise motion-speed control, ensure long-range coherence and detail preservation during rapid movements. Our experiments demonstrate a breakthrough beyond the 20-second limit, producing stable 720p/30fps videos exceeding one minute while showcasing versatility across five dance genres conditioned on audio and text. This work sets a new state-of-the-art in coherent, long-form music-to-dance synthesis.

\subsubsection{Limitations and Future Work:}

Three key areas need further exploration: 

\begin{itemize}
    \item Identity Consistency: We plan to integrate face-embedding constraints to maintain facial consistency across long sequences.
    \item Semantic Alignment: Future efforts will employ multimodal semantic encoders to enhance contextual expressiveness in the generated dances.
    \item Multi-Dancer Scenarios: We aim to extend our architecture to model inter-dancer interactions for synchronized group choreography.
\end{itemize}

\subsubsection{Ethics Statement:} Our research focuses on advancing human image animation techniques and disavows malicious uses, like deceptive deepfakes. All synthesized outputs will be clearly labeled as artificially generated.

%
%
\bibliographystyle{splncs04}
\bibliography{main}

\end{document}